\documentclass[default]{sn-jnl}

\usepackage{graphicx}%
\usepackage{multirow}%
\usepackage{amsmath,amssymb,amsfonts}%
\usepackage{amsthm}%
\usepackage{mathrsfs}%
\usepackage[title]{appendix}%
\usepackage{xcolor}%
\usepackage{textcomp}%
\usepackage{manyfoot}%
\usepackage{booktabs}%
\usepackage{algorithm}%
\usepackage{algorithmicx}%
\usepackage{algpseudocode}%
\usepackage{listings}%
\usepackage{longtable}
\usepackage{orcidlink}

\begin{document}

\title[Article Title]{Generative Artificial Intelligence: A Systematic
Review and Applications}

\author*[1]{\fnm{Sandeep Singh} \sur{Sengar}\orcidlink{0000-0003-2171-9332}}\email{SSSengar@cardiffmet.ac.uk, Phone: +44-7780390651}

\author[1]{\fnm{Affan Bin} \sur{Hasan}}\email{ahasan@cardiffmet.ac.uk}

\author[2]{\fnm{Sanjay} \sur{Kumar}}\email{sanjay.kumar@dtu.ac.in}

\author[1]{\fnm{Fiona} \sur{Carroll }}\email{fcarroll@cardiffmet.ac.uk}

\affil[1]{\orgdiv{Cardiff School of Technologies}, \orgname{Cardiff Metropolitan University}, \orgaddress{ \city{Cardiff}, \postcode{CF5 2YB}, \country{United Kingdom}}}

\affil[2]{\orgdiv{Department of Computer Science and Engineering}, \orgname{Delhi Technological University}, \orgaddress{\city{New Delhi}, \postcode{110042}, \country{India}}}

\abstract{In recent years, the study of artificial intelligence (AI) has undergone a paradigm shift. This has been propelled by the groundbreaking capabilities of generative models both in supervised and unsupervised learning scenarios. Generative AI has shown state-of-the-art performance in solving perplexing real-world conundrums in fields such as image translation, medical diagnostics, textual imagery fusion, natural language processing, and beyond. This paper documents the systematic review and analysis of recent advancements and techniques in Generative AI with a detailed discussion of their applications including application-specific models. Indeed, the major impact that generative AI has made to date, has been in language generation with the development of large language models, in the field of image translation and several other interdisciplinary applications of generative AI. Moreover, the primary contribution of this paper lies in its coherent synthesis of the latest advancements in these areas, seamlessly weaving together contemporary breakthroughs in the field.  Particularly, how it shares an exploration of the future trajectory for generative AI. In conclusion, the paper ends with a discussion of Responsible AI principles, and the necessary ethical considerations for the sustainability and growth of these generative models.}


\maketitle

\section{Introduction}\label{sec1} The recent advancement in Artificial Intelligence has been mainly the result of Generative Artificial Intelligence (often referred to as Generative AI or GenAI) being introduced. Generative AI  encompasses artificial intelligence systems with the ability to create text, images, or various forms of media through the utilization of generative models. These models acquire an understanding of the underlying patterns and structures within their training data, subsequently producing fresh data that share similar traits and characteristics. The motivation of this systematic review is to gather, evaluate, and synthesize existing research on a GenAI. This paper presents a systematic review that highlights key applications and variations of the architecture of Generative Artificial Intelligence models and their performance. We conducted this review to (a) understand the state-of-the-art generative AI techniques including summarizing key methodologies, algorithms, and findings across a range of studies (b) systematically review a large body of literature, which includes emerging trends, common challenges, and recurring patterns in the development and application of generative AI techniques (c) compare and contrast different generative AI approaches, such as Autoencoders, Generative Adversarial Networks, Transformers, and Diffusion models (d) explore successful applications of generative AI such as image translation, video synthesis and generation, natural language processing, knowledge graph generation, etc. (e) identify ethical challenges and propose solutions for responsible AI development.

In this research, we outline the most recent research and advancement in the field of Generative Artificial Intelligence. It details the approach used to navigate and analyze cutting-edge developments, ensuring a comprehensive and insightful review of the current landscape in Generative AI. The following criteria were applied for searching the used research papers.\\
\textbf{Time Period:} This paper presents a comprehensive overview of the advancements and applications of Generative AI, focusing on significant developments between 2018 and 2023. Additionally, it offers a concise historical perspective, tracing the evolution of foundational models from 2012 to 2018, which laid the groundwork for the current state of Generative AI techniques. This historical context enriches the understanding of the field's rapid progression and its burgeoning applications.\\
\textbf{Keywords:} This paper employs a targeted keyword search strategy, incorporating specific terms such as `Generative Adversarial Networks', `Transformers', `Variational Autoencoders', and `Diffusion Models'. This approach also includes searching for advancements in `image translation', `video synthesis', and various applications of Generative AI in `natural language processing' and `knowledge graph generation'. This methodology ensures a focused and comprehensive review of the latest developments in the field of Generative AI.\\
\textbf{Databases: }The work primarily sources relevant literature from Google Scholar, focusing on the specified timeframe. It selectively includes research that showcases advancements in generative models. This criteria ensures the inclusion of studies where developed models were rigorously tested on well-recognized datasets, and where results are communicated effectively and clearly. This approach guarantees that the paper presents a detailed and credible overview of significant developments in the field of Generative AI. \\
\textbf{Inclusion Criteria: } This work exclusively incorporates peer-reviewed papers, conference, and journal papers that are written in English. It emphasizes studies that highlight either significant advancements or innovative applications in the realm of Generative AI, ensuring that the focus remains on cutting-edge and impactful developments within this field.\\
\textbf{Exclusion Criteria: }This paper meticulously filters its sources, excluding non-peer-reviewed materials, papers not written in English, and studies that fall outside the 2012-2023 timeframe. Additionally, it deliberately omits any papers that do not directly contribute to the advancement or understanding of Generative AI, ensuring a focused and relevant academic discourse.\\
\textbf{Evaluation Criteria: }In each subsection evaluating the advancements in Generative AI techniques, the paper compares the performance of various models using standardized datasets commonly cited in the field. This comparison focuses on how different state-of-the-art models perform on these datasets, providing a clear and consistent basis for assessing the progress and effectiveness of these techniques in their respective domains.

The following is the summary of the major contributions of our work- 

\begin{itemize}
    \item \textbf{Paradigm Shift in Artificial Intelligence:} The paper discusses the paradigm shift in artificial intelligence and highlights the significant impact of generative models in the field of machine learning. 
    \item \textbf{Historical Context:} The paper includes a section that gives a straightforward overview of how key AI models have developed from 2012 to 2018, helping to better understand how the field has grown and changed over time. 
    \item \textbf{Real-World Uses of Generative AI:} The paper describes how Generative AI is used in different areas like image translation, diagnosing medical conditions, combining text and images, processing natural language, etc.
    \item \textbf{Systematic Review of Generative AI: }The work provides a comprehensive review and analysis of recent advancements in Generative AI, focusing on techniques and applications, including application-specific models. We have also provided information on relevant datasets for each used application. 
    \item \textbf{Impact on Language and Image Translation: } The paper discusses the major impact of generative AI in language generation with large language models and in the field of image translation. 
    \item \textbf{Responsible AI Principles:} The paper ends with a discussion on Responsible AI principles and ethical considerations necessary for the sustainability and growth of generative models. 

\end{itemize}

After this introductory section, Sections~\ref{sec2} and~\ref{sec3} review the basic early architecture of Generative adversarial Networks and their variants. Section~\ref{sec4} deeply explores the recent applications and the advancements in Generative AI application-specific techniques. Section~\ref{sec5} provides the  Challenges and opportunities of Generative AI. Lastly, Section~\ref{sec6} concludes the work and highlights the future directions of generative AI.

\section{What is Generative Artificial Intelligence? }\label{sec2}
As discussed Generative Artificial Intelligence refers to artificial intelligence systems with the capability to create text, images, or other forms of media through the utilization of generative models. These models acquire an understanding of patterns and structures within their training data, subsequently generating novel data with akin characteristics. Generative Artificial Intelligence encompasses various types, each tailored for specific tasks or forms of media generation. The following are some of the more well-known types: Generative Adversarial Networks (GANs)~\cite{boroujeni2024ic}, Transformer-based Models (TRMs)~\cite{reza2022multi}, Variational Autoencoders (VAEs)~\cite{fei2023novel}, and Diffusion models (DMs)~\cite{shao2023data}, to name a few. The following sections will discuss these in more detail.

\subsection{Generative Adversarial Networks}\label{subsec1}

A generative adversarial network (GAN) is a class of machine learning framework and a prominent framework for approaching generative AI. The aspect that is novel in this generative adversarial network set-up is that it does not depend upon heavily annotated training data. Moreover, the architecture that it affords is quite unique from the conventional Deep Neural Networks~\cite{ganreview}.
 Indeed, it consists of two major components named \emph{Generator} and \emph{Discriminator}. The main operation of the generator is to keep on generating the fake data using the noise while the purpose of the discriminator is to distinguish whether the generated image is real or fake. The discriminator is trained using the real images of the domain that the generator is trying to synthetically produce and the discriminator’s sole purpose is to identify whether the output produced by the generator is fake or not. The overall system is based on the zero-sum game dynamics, the winner will remain unchanged and the loser model each time has to modify its parameters, it will keep on doing this until the discriminator is unable to detect whether the generator output is fake or not~\cite{goodfellow2014generative}. The sole purpose of this is to build a powerful generator model that generates synthetic data that looks real. 
 
\begin{figure}[h]
\centering
\includegraphics[width= 11cm]{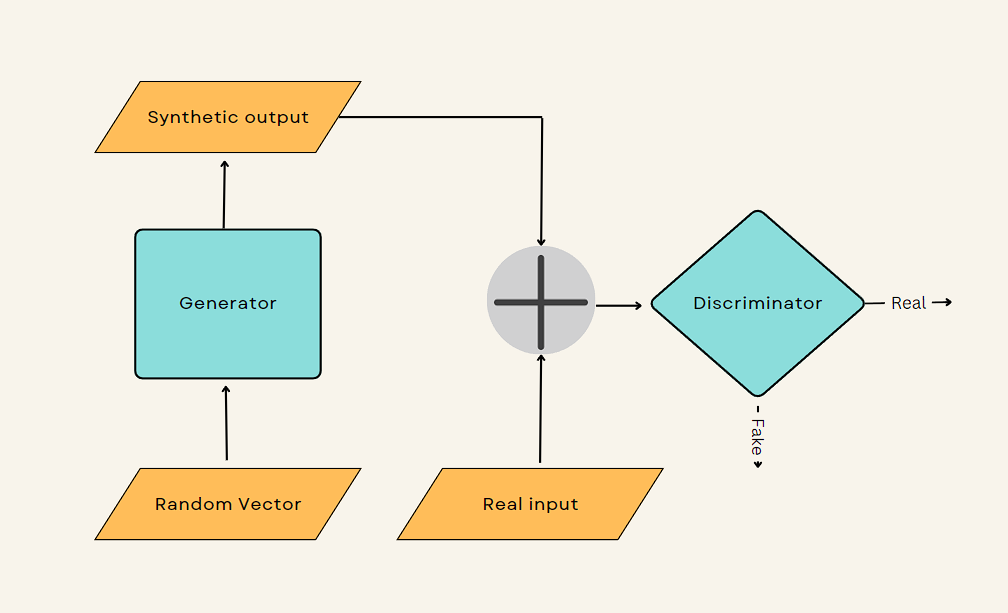}
  \caption{GAN- Generator and discriminator working}\label{1}
\end{figure}

 Fig.~\ref{1} demonstrates how the generator and discriminator work together. The generator aims to deceive the discriminator by providing the synthetically generated image with the objective that it is proven real. The discriminator discerns between genuine and counterfeit images and generates the output signal. This output signal then goes to both the generator and discriminator, allowing the generator to produce better synthetic output. And, in case the discriminator fails to prove the image is fake, it also uses the signal to change its weights to give better predictions. In this entire architecture, it is important to note that only the discriminator has access to the real image, synthetic image, and its own signal output while the generator only learns from the output signal of the discriminator \cite{ganreview}.  

During the initial stages of development, Generative Adversarial Network (GAN)-based models encountered significant challenges in their training process. These difficulties primarily revolved around issues like training divergence and model collapse \cite{mescheder2018traininggans}. Training divergence refers to situations where the GAN's generator and discriminator fail to achieve a stable equilibrium during training, leading to oscillations and unreliable model outputs. This problem results in inconsistent and sub-optimal generation performance, hindering the GAN's ability to produce high-quality samples. On the other hand, model collapse occurs when the GAN's generator produces limited and repetitive outputs, ignoring a large portion of the data distribution. This phenomenon causes the generator to focus on a small subset of data points, resulting in a lack of diversity and novelty in the generated samples.
Addressing these challenges has been a main focus in the advancement of GAN-based models, with numerous research efforts aimed at improving stability, convergence, and diversity during the training process. As a result, substantial progress has been made, leading to the development of more robust and effective GAN architectures, which have significantly enhanced the performance and applicability of these generative models.

\subsection{Transformers}\label{subsec2}
Further Generative AI techniques called \emph{Transformers} were introduced by Vaswani et al.~\cite{NIPS2017_attention}. This breakthrough architecture laid the foundation for various tasks, including machine translation and language generation, and it continues to influence subsequent neural network designs. The paper's emphasis on attention mechanisms highlighted their pivotal role in sequence-to-sequence tasks, advancing the state of the art. Transformers use both the self-attention and Multi-Head Attention mechanisms to learn the dependencies between the objects regardless of the distance between them and to learn the different relations and patterns between the input respectively. Often in Natural Language Processing, these methods are combined with positional encoding added to the input sequence to make the transformers keep track of the position of a specific word in an input sequence. Transformers are commonly used to build Generative AI Models such as Generative Pre-trained Transformers (GPT) models which are capable of generating coherent and contextually relevant text \cite{radford2018improving}. Bidirectional Encoder Representations from Transformers (BERT) and Open AI GPT are based on transformers.

\subsection{Variational Autoencoders}\label{subsec3}
Another model in the field of generative AI is Variational Autoencoders (VAEs), introduced by Kingma et al.~\cite{kingma2022autoencoding}. As the name suggests the VAEs consist of an encoder and a decoder. The purpose of an encoder is to encode the given input in a lower dimension called \emph{latent space} and the decoder decodes that latent output of the encoder into its original input shape. During this whole process, variation is introduced to the latent space by using the standard Gaussian distribution. The main goal is to achieve the output with a similar mean and variance as the given input after the introduction of the variance. This provides a structured way to learn meaningful representations of data and then generate new samples from that data distribution.

\subsection{Diffusion Models}\label{subsec4}
The Diffusion Models have been designed to improve the performance of the Simple Generative Adversarial Network, the technique was introduced by Salimans et al ~\cite{NIPS2016_diffusion}. At a later stage, Kingma et al ~\cite{NIPS2016_diff} introduced a variant of the diffusion model called Inverse Autoregressive Flow (IAF) as a building block for generative models. IAF is a type of normalizing flow. This is a type of generative model that aims to learn complex probability distributions by transforming a simple base distribution into the target distribution through a series of invertible transformations.

\section{Evolution of Generative AI Models: A Look at Earlier Variants}\label{sec3}
\subsection{Earlier GAN Variants}\label{subsec3.1}

In the early stages of the introduction of Generative Artificial Intelligence models the major issue that researchers were facing was the convergence problem of generative models~\cite{mescheder2018traininggans}. To avoid this problem, different approaches were adopted by the researchers to make the GAN more stable (e.g., by understanding the behavior of GAN training). In detail, Mescheder et al.~\cite{mescheder2018numerics} explained the analysis of local convergence and stability properties during the training of GAN. This involves an examination of the eigenvalues of the Jacobian matrix associated with the gradient vector field. Specifically, when the equilibrium point is characterized by solely negative real-part eigenvalues in the Jacobian, GAN training demonstrates local convergence, specifically when utilizing relatively small learning rates. However, the situation changes when the eigenvalues of the Jacobian are situated on the imaginary axis. In such cases, the local convergence of GAN training is generally compromised. Also, it is important to note that if the eigenvalues are in proximity and not directly on the imaginary axis, the training algorithm may necessitate exceedingly small learning rates to achieve convergence~\cite{mescheder2018numerics}.

The study by Mescheder et al.~\cite{mescheder2018numerics} identified instances of eigenvalues near the imaginary axis in practical scenarios. This observation does not definitively address whether such proximity to the imaginary axis is a prevalent phenomenon. Furthermore, it does not conclusively establish whether these eigenvalues are the fundamental cause behind the training instabilities that practitioners commonly encounter in their GAN training endeavors. Moreover, Nagarajan et al.~\cite{NIPS2017_7e0a0209} contributed a partial response to this query. They demonstrated that in the context of absolutely continuous data and generator distributions, these findings establish that GANs exhibit local convergence for sufficiently small learning rates. However, to emphasize, this assertion relies on the premise of absolute continuity.

Goodfellow et al.~\cite{goodfellow2014generative} presented the basic GAN architecture, other researchers also advanced more variants with some architectural differences. However, due to the potentially limiting overlap between real and generated data distributions, the Jensen-Shannon divergence presented in the objective function can become a constant value. It is this phenomenon that contributes to the challenge of the vanishing gradient, hindering effective training of GANs when employing gradient descent methods. To address the vanishing gradient problem, the Wasserstein GAN (W-GAN) was introduced \cite{wassersteingan17}, using the EarthMover distance instead of the Jensen-Shannon divergence to compare real and generated data distributions. W-GAN employs a critic function 'f' with a Lipschitz constraint as its discriminator, significantly improving GAN training stability. However, W-GAN may still face issues like suboptimal sample generation and occasional convergence problems in specific cases. In order to restrict the discriminative capacity of the discriminator, an alternative approach has also been introduced by \cite{qi2018losssensitive} in the form of Loss-Sensitive GAN (LS-GAN). Both W-GAN and LS-GAN retain the fundamental GAN architecture.

\begin{figure}[H]
\centering
\includegraphics[width= 9cm, height=16cm]{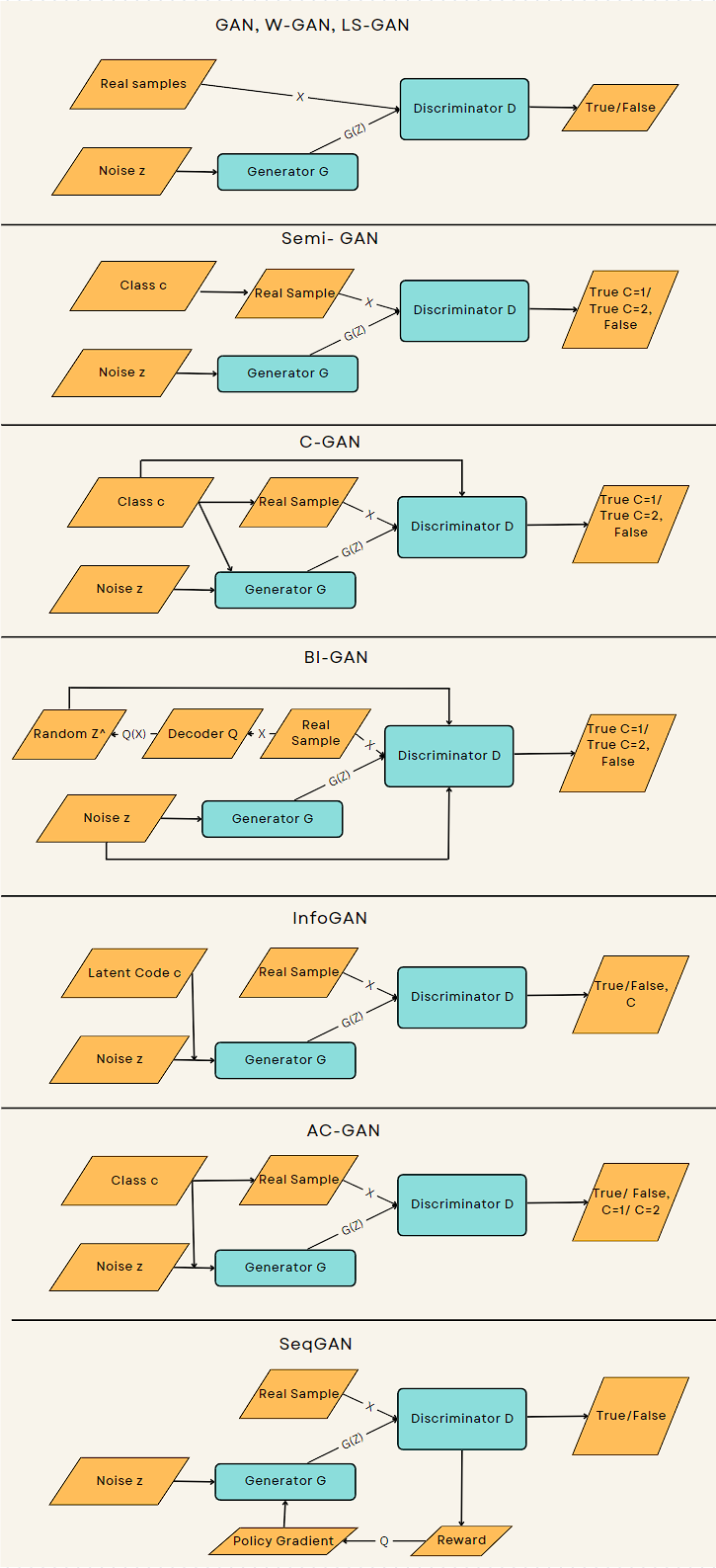}
  \caption{Basic GANS Variants}\label{2}
\end{figure}

Moreover, Qi et al.~\cite{odena2016semisupervised} introduced the Semi-GAN model, which involves the incorporation of real data labels into the discriminator's training process. Additionally, semi-GAN is an approach involving the integration of auxiliary information 'y' into both the generator 'G,' the discriminator 'D,' and the real data 'x' for the discriminator \cite{mirza2014conditional}. This auxiliary information can encompass labels or other supplementary data. In the context of conventional GANs, the primary objective revolves around acquiring a generative model capable of mapping latent variable distributions to intricate real data distributions. Expanding upon this concept,  Donahue et al. \cite{donahue2016adversarial} propose Bidirectional GANs (BiGANs) to facilitate the mapping of real data to the latent variable space, thereby enabling feature learning. BiGANs extend the fundamental GAN structure by incorporating an additional decoder 'Q,' which facilitates the transformation of real data 'x' into the latent space. Consequently, this modification transforms the optimization problem into the form \(\min_{G, Q} \max_D f(D, Q, G)\).

Chen et al. \cite{chen2016infogan} introduce InfoGAN to capture mutual information between a subset of latent variables and observed data.

In InfoGAN, the correlation is quantified using $I(c; G(z, c))$, where $c$ is a latent code and $G(z, c)$ is the generated output. The objective function is:
\[
\min_G \max_D \{ f(D, G) - \lambda I(c; G(z, c)) \}
\]
Here, $f(D, G)$ includes the adversarial term and penalty term $\lambda I(c; G(z, c))$. The goal is to minimize the generator’s loss while maximizing the discriminator’s loss with respect to mutual information.

Due to the challenge of computing $p(c|x)$, a lower bound estimation is used through variational information maximization.







Odena et al. \cite{odena2017conditional} introduce the auxiliary classifier GAN (AC-GAN) approach tailored for semi-supervised synthesis. Their formulated objective function comprises two integral components: the logarithmic likelihood related to the accurate data source and the corresponding accurate class. The essence of AC-GAN lies in its capacity to seamlessly integrate label information into the generator and to adapt the discriminator's objective function accordingly. This integration yields noticeable enhancements in the generative and discriminative capabilities of the GAN framework.

In another context, Yu et al. \cite{yu2017seqgan} introduced SeqGAN, a pioneering framework for sequence generation using GANs. It extends GANs to handle discrete token sequences, treating the generator as a stochastic policy in reinforcement learning. SeqGAN employs policy gradient-based mechanisms to enhance sequence generation by effectively propagating errors from the discriminator. These advancements build upon the foundational work of GANs \cite{goodfellow2014generative}.


\subsection{Earlier Transformer Variants}\label{subsec3.2}
The concept of Transformers was introduced by Vaswani et al. \cite{NIPS2017_attention}. It was a revolutionary step in the field of generative AI specifically in natural language processing and generating synthetic content. The basic concept of the Transformers was introduced in \cite{NIPS2014_trans} by Sutskever et al as a sequence modelling technique. The basic technique of pretraining transformers was introduced and used as a state-of-the-art technique by \cite{qiu2020pre}. This was used in answering different queries and also used as a chatbot to give results that are competitive and accurate. Indeed, these early Transformer Developments paved the way for State-of-the-Art NLP Chatbots.

\subsection{Earlier Variational Autoencoder Variants}\label{subsec3.3}
Variational Autoencoders \cite{kumar2023community} is one of the oldest techniques of unsupervised learning and generative modeling. The foundational work of Variational Autoencoders was done by Kingma at el in \cite{kingma2013auto}. The Variational Autoencoders model combines the probabilistic modeling with the basics of Autoencoders. This concept not only learns the properties of the latent space but they also learn the probabilistic distribution of it, which gives them the ability to generate new synthetic data samples. Early variants of autoencoders also include denoising autoencoders which use denoising techniques whilst trained locally to get rid of corrupted versions of their inputs \cite{vincent2010stacked}. Moreover, \cite{masci2011stacked} used the auto-encoders coupled with the convolutional network to solve the image recognitional problems. Without a doubt, the pioneering work of \cite{kingma2013auto} has paved the way for numerous subsequent developments and applications of VAEs in a wide range of domains, including image generation, natural language processing, and more.

\subsection{Earlier Diffusion Model Variants}\label{subsec3.4}
Diffusion-based models employ a sequential diffusion process to iteratively transform simple data distributions into complex, high-dimensional ones.  The Non-Linear independent component estimation (NICE) introduced the concept of invertible transformations as a foundation for generative artificial intelligence~\cite{dinh2015nice}. This was followed by Real NVP (Real Non-Volume Preserving), which expanded the capabilities by incorporating neural networks into the transformation process~\cite{dinh2017density}. Additionally, Glow (Generative Latent Optimization) extended these ideas to high-resolution image generation, highlighting the potential of diffusion-based models in computer vision \cite{kingma2018glow}. Furthermore, Diffusion Probabilistic Models (DPMs) leveraged the diffusion process to model the likelihood of data samples, making it an essential contribution to the development of diffusion-based generative models. Continuous-time flows (CTFs) diffusion models ventured into continuous-time modelling using stochastic differential equations \cite{grathwohl2018ffjord}. These earlier works have laid the foundation for an exciting and rapidly evolving field of generative modelling using diffusion-based techniques.

\section{Advancements in Generative AI and Their Diverse Applications}\label{sec4}
\subsection{Generative AI for Image Translation}\label{subsec4.1}
Image translation~\cite{goodfellow2014generative} is becoming a rapidly growing technology, particularly within the realm of medical applications. This innovation holds remarkable potential, not only in terms of cost-saving implications related to equipment usage but also in the facilitation of informed medical decisions.

The performance of generative AI models, particularly in the subfield of image translation, is typically assessed using specialized datasets. Among these, two notable datasets stand out: ImageNet \cite{yang2021imagenetfaces}, ClebA \cite{liu2015faceattributes}, and in the field of medical science: MIMIC \cite{johnson2016mimic}, BRATS \cite{menze2014multimodal}, FastMRI \cite{tibrewala2023fastmri} and ChestX-ray \cite{wang2017chestx} Each of these datasets is uniquely designed to challenge and evaluate the models' abilities to accurately and effectively translate images, providing a comprehensive benchmark for their performance capabilities. These datasets are freely accessible to researchers for testing their models, under certain conditions. Users must properly cite the source of the dataset in their work. For datasets containing medical data, researchers are required to sign a Data Use Agreement. This agreement sets forth guidelines on the appropriate usage and security of the data and strictly prohibits any attempts to identify individual patients. This ensures that while fostering innovation and research, the datasets are used ethically and responsibly.

\begin{figure}[H]
    \centering
    \includegraphics[width=0.7\textwidth]{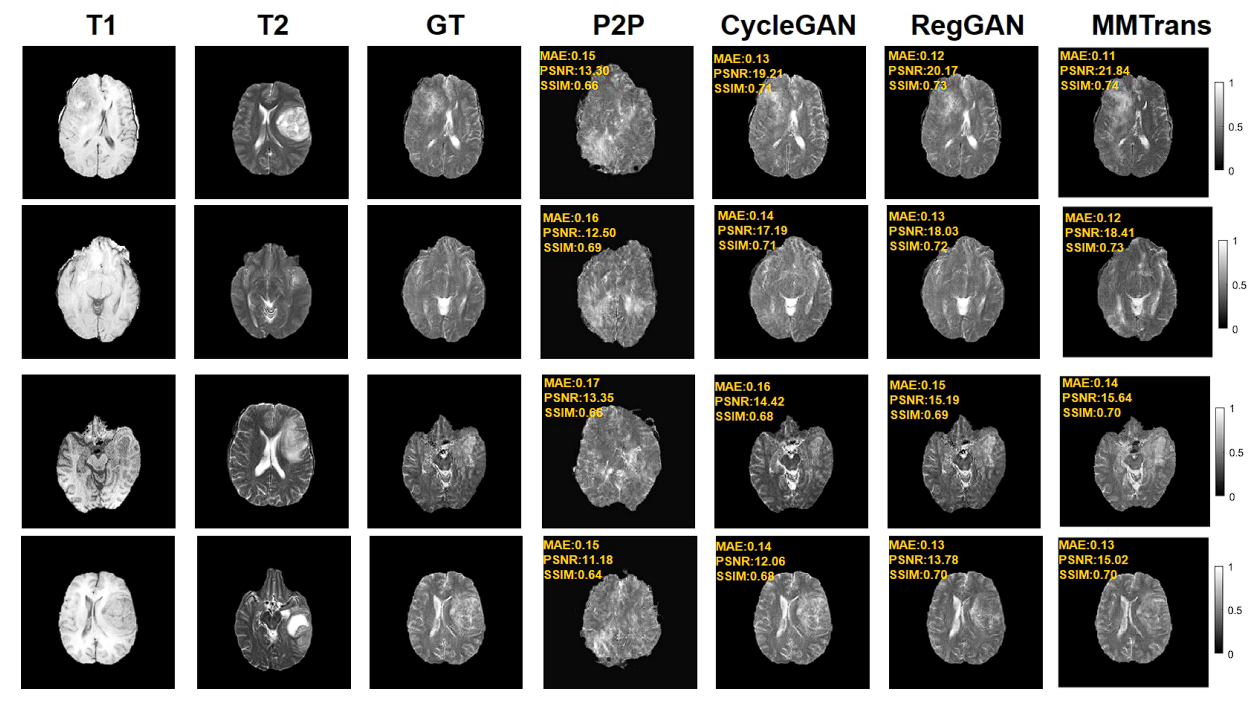}
     \caption{qualitative outcomes obtained from various translation techniques employed to generate T2 images from T1 images within the unpaired BraTs2018 dataset, Source: \cite{yan2022swin}}
     \label{swin}
\end{figure}

The utilization of AI-driven image translation yields images that are not only more polished and precise but also empower medical professionals with a heightened ability to discern critical information \cite{yan2022swin}. Yan et al. proposed a GANs-based model that uses the \emph{Swin Transformers} in the Generator. The Swin Transformer represents a notable stride forward in the evolution of architecture. Its most remarkable enhancement entails the replacement of the conventional multiple self-attention (MSA) modules with an innovative shift window-based module while keeping the remaining layers largely unchanged. This transformer-based generator allows for the production of the output content which is the same as source images and the same information required by the target image. They tested the model using the BraTs2018 \cite{menze2014multimodal} and FastMRI \cite{sengar2023multi} datasets.
The Swin-based Transformers method attains its highest level of performance in the specific task of converting T1 mode to T2 mode images using the clinical brain MRI dataset. Moreover, they conducted evaluations using the unpaired \emph{BraTs2018} dataset (see the results depicted in the Figure \ref{swin}). These highlight that the innovative MMTrans approach stands out as the leader in terms of translation performance.

Indeed, Figure \ref{swin} shows that the introduced MMTrans method performs better in translating the image when compared to other methods such as Pix2Pix \cite{isola2018imagetoimagepix2pix}, CycleGAN \cite{cyclic2017} and RegGAN \cite{rezagholiradeh2018reg}. The Image of MMTrans is closest to the Ground Truth (GT).

Furthermore, Dar et al. \cite{dar2018image} used Conditional GANs to solve the problem of Image translation in MRI. This variant of GANs was introduced for image-image translation with the conditional label given to both generator and discriminator to instruct about what they have to forge and to predict real or fake respectively \cite{isola2018imagetoimagepix2pix}. In fact, there are two types of GAN variants present for the purpose of image translation, Pix2Pix GAN is a conditional GAN~\cite{isola2018imagetoimagepix2pix}, where the generator takes both an input image and a target condition as input and then generates an output image that adheres to the specified condition.  However, for these types of GANs pixel-aligned images are required which is quite difficult to acquire~\cite{yan2022swin}. Moreover, unpaired GANs do not require corresponding pairs of images for training. Instead, they focus on learning the mapping between two domains by using cycle consistency as a constraint~\cite{torbunov2023uvcgan}.

\begin{figure}[H]
    \centering
    \includegraphics[width=0.7\textwidth]{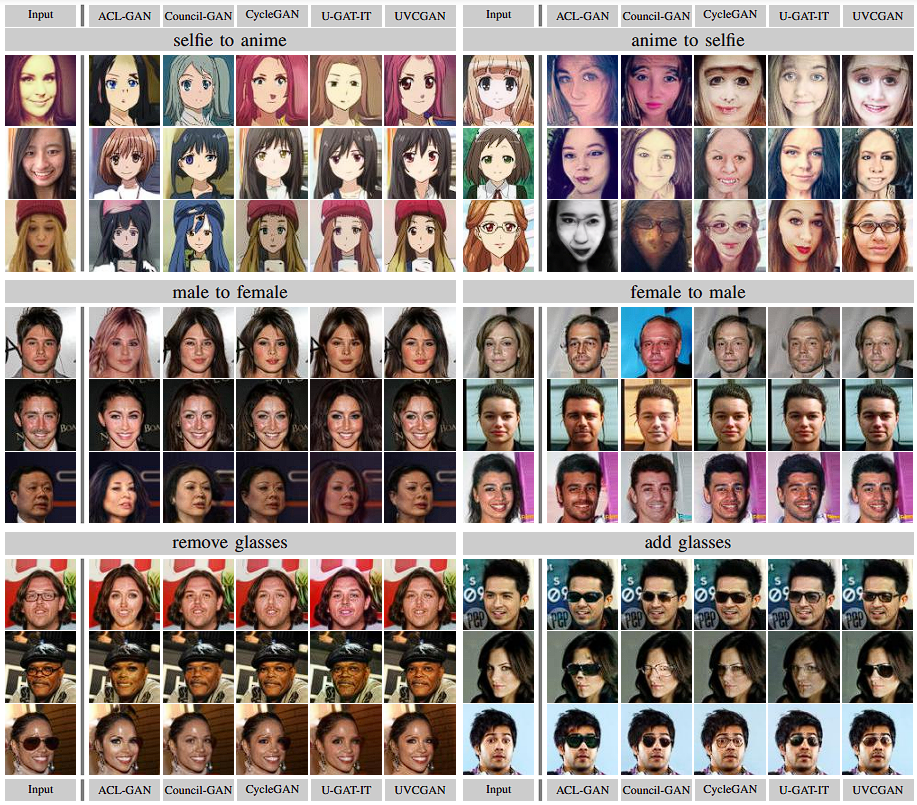}
     \caption{unpaired UVCGAN vs Others image-to-image translation, Source: \cite{torbunov2023uvcgan}}
     \label{uvcgan}
\end{figure}

Indeed, unpaired image translation presents a significant challenge. The objective is to ensure that translating an image from one domain to another should not compromise its fundamental characteristics (e.g., allowing for a seamless reversion back to its initial state)\cite{cyclic2017}. To address this issue, the Cyclic Generative Adversarial Network (Cyclic GAN) was developed \cite{cyclic2017}. Torbunov et al. \cite{torbunov2023uvcgan} developed a Vision Transformer based GAN (UVCGAN). This works on the principle of cyclic GANs and without sacrificing the image regeneration capabilities gives better results than previous simple cyclic models. The evaluation of image-to-image translation performance commonly employs two widely accepted metrics, namely Frechet Inception Distance (FID) \cite{NIPS2017_FID} and Kernel Inception Distance (KID) \cite{binkowski2018demystifying}. These metrics quantify the similarity between the translated images and those within the target domain, with a lower score indicative of higher similarity. UVCGAN model's superior performance is evident across most image-to-image translation tasks, as illustrated in Table \ref{1}. Operating similarly to a CycleGAN-like model, their approach consistently produces translated images that exhibit strong correlations with the input images, capturing essential aspects like hair color and facial orientations (as exemplified in Figure \ref{uvcgan}).

\begin{table}
  \centering
  \caption{FID and KID scores of UVCGAN and other models. Lower is better, Source:\cite{torbunov2023uvcgan}}
  \begin{tabular}{|l|c|c|c|c|}
    \hline
    \multirow{2}{*}{Model} & \multicolumn{2}{c|}{Selfie to Anime} & \multicolumn{2}{c|}{Anime to Selfie} \\
    & FID & KID ($\times100$) & FID & KID ($\times100$) \\
    \hline
    ACL-GAN & 99.3 & 3.22 $\pm$ 0.26 & 128.6 & 3.49 $\pm$ 0.33 \\
    Council-GAN & 91.9 & 2.74 $\pm$ 0.26 & 126.0 & 2.57 $\pm$ 0.32 \\
    CycleGAN & 92.1 & 2.72 $\pm$ 0.29 & 127.5 & 2.52 $\pm$ 0.34 \\
    U-GAT-IT & 95.8 & 2.74 $\pm$ 0.31 & 108.8 & 1.48 $\pm$ 0.34 \\
    UVCGAN & 79.0 & 1.35 $\pm$ 0.20 & 122.8 & 2.33 $\pm$ 0.38 \\
    \hline
    \multirow{2}{*}{Model} & \multicolumn{2}{c|}{Male to Female} & \multicolumn{2}{c|}{Female to Male} \\
    & FID & KID ($\times100$) & FID & KID ($\times100$) \\
    \hline
    ACL-GAN & 9.4 & 0.58 $\pm$ 0.06 & 19.1 & 1.38 $\pm$ 0.09 \\
    Council-GAN & 10.4 & 0.74 $\pm$ 0.08 & 24.1 & 1.79 $\pm$ 0.10 \\
    CycleGAN & 15.2 & 1.29 $\pm$ 0.11 & 22.2 & 1.74 $\pm$ 0.11 \\
    U-GAT-IT & 24.1 & 2.20 $\pm$ 0.12 & 15.5 & 0.94 $\pm$ 0.07 \\
    UVCGAN & 9.6 & 0.68 $\pm$ 0.07 & 13.9 & 0.91 $\pm$ 0.08 \\
    \hline
    \multirow{2}{*}{Model} & \multicolumn{2}{c|}{Remove Glasses} & \multicolumn{2}{c|}{Add Glasses} \\
    & FID & KID ($\times100$) & FID & KID ($\times100$) \\
    \hline
    ACL-GAN & 16.7 & 0.70 $\pm$ 0.06 & 20.1 & 1.35 $\pm$ 0.14 \\
    Council-GAN & 37.2 & 3.67 $\pm$ 0.22 & 19.5 & 1.33 $\pm$ 0.13 \\
    CycleGAN & 24.2 & 1.87 $\pm$ 0.17 & 19.8 & 1.36 $\pm$ 0.12 \\
    U-GAT-IT & 23.3 & 1.69 $\pm$ 0.14 & 19.0 & 1.08 $\pm$ 0.10 \\
    UVCGAN & 14.4 & 0.68 $\pm$ 0.10 & 13.6 & 0.60 $\pm$ 0.08 \\
    \hline
  \end{tabular}
  
  \label{table:1}
\end{table}

The high-quality result produced by UVCGAN in Figure \ref{uvcgan} is paramount for enhancing scientific simulations. It can be observed that translations generated by ACL-GAN and Council-GAN tend to overly emphasize features that aren't pivotal for achieving the intended translation, such as non-essential attributes like hair color, background color, and length. Even in some cases, Council-GAN changed the background.
Another application under the umbrella of image translation is the Synthetic Aperture Radar (SAR) image translation \cite{zhao2022comparative}. Both the techniques of paired and unpaired GANs are used for this application. Indeed, the SAR-generated images are not that visually clear but they can be captured at any time so it's a better technique when compared to optical imaging. Particularly, if it is combined with the GANs image translation methods to convert the SAR-generated image into an optical high-resolution clear image \cite{zhao2022comparative}. Different GANs methods of unpaired image translation such as CycleGAN \cite{cyclic2017}, NICE GANs \cite{niceunsupervised}, Attn-CycleGAN \cite{Lin_2021_WACV} and paired GANs such as Pix2Pix \cite{isola2018imagetoimagepix2pix}, and Bicycle GANs \cite{zhu2017toward} are used for Satellite image translation. In Wei et al. \cite{wei2023cfrwd} the researchers presented a new technique using the generative AI that is specifically designed for the translation of unpaired SAR images to optical images. In detail, they introduced an approach known as Cross-Fusion Reasoning and Wavelet Decomposition GAN (CFRWD-GAN)~\cite{wei2023cfrwd}. The primary objective of CFRWD-GAN is twofold: to effectively retain structural intricacies and elevate the quality of high-frequency band details. This is achieved through a unique framework that integrates cross-fusion reasoning (CFR) structure, adept at preserving both high-resolution, fine-grained features and low-resolution semantic attributes throughout the entire process of feature reasoning. Additionally, to address speckle noise inherent in SAR images, the method employs discrete wavelet decomposition (WD), enabling the translation of high-frequency components. Through the convergence of these techniques, CFRWD-GAN demonstrates its capability to significantly enhance the translation process for unpaired image-to-image scenarios. The model was evaluated using Root Mean Squared Error (RMSE)~\cite{jain2023age,sengar2022content}, structural similarity index (SSIM) \cite{imagequality,lakshmi2023classification}, peak signal-to-noise ratio (PSNR) \cite{TANCHENKO2014874psnr,sengar2020motion}, learned perceptual image patch similarity (LPIPS) \cite{zhang2018unreasonable,sengar2016moving} and produced a better result than the other state of art models.
 \begin{figure}[H]
    \centering
    \includegraphics[width=0.7\textwidth]{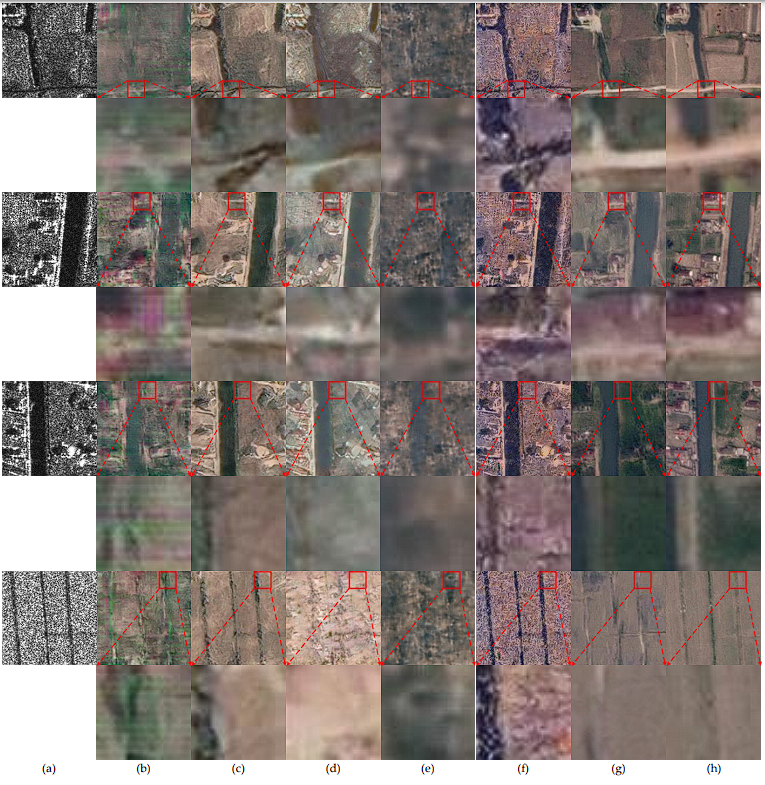}
     \caption{Highlighted with red boxes and magnified for emphasis, the images are presented in the following order: (a) SAR images, (b) Pix2Pix, (c) CycleGAN, (d) S-cycle-GAN, (e) NICE-GAN, (f) GANILLA, (g) CFRWD-GAN, and (h) ground truth optical images., Source: \cite{wei2023cfrwd}}
     \label{sar}
\end{figure}

The other technique for image generation and translation is Variational Autoencoders developed by Kingma et al.~\cite{kingma2013auto}.  Furthermore,  Zhu et al. \cite{NIPS2017_cve}
compared the generative capabilities of Conditional Variation Autoencoders and also compared it with the other generative techniques for image generation and translation. The main idea of this work was to find the algorithm that performs the best in balancing the diversity and realism in the generated data.  The best-performing model in their work was the Bicycle GANs. 
 \begin{table}[h!]
  \centering
  \caption{Generative AI and Its Applications in Image Translation}
  \label{table:2}
  \begin{tabular}{p{3.5cm}p{3.5cm}p{4.5cm}}
    \toprule
    Domain & Methods & References \\
    \midrule
    Medical-MRI & MM-Transformers, Cyclic-GAN, Pix2Pix GAN, VAE & \cite{yan2022swin, dar2018image, ali2022role, dimitriadis2022enhancing, ahmad2022brain, xu2023generative, alzheimer22} \\
    Satellite Image Translation & Cyclic-GAN, Pix2Pix GAN, NICE-GAN, Attn-CycleGAN, PSGAN & \cite{zhao2022comparative, paola2023correction, liu2020psgan} \\
    Facial Expression Editing & VAE, UPGPT & \cite{yeh2016semantic, cheong2023upgpt, fontanini2023automatic} \\
    Style Transfer & VAE, GANs, DD-GAN & \cite{zuo2022style, tahir2022diverse, Atapour-Abarghouei_2018_CVPR} \\
    Text-to-Image Translation & TextControlGAN & \cite{ku2023textcontrolgan} \\
    Image Upscaling & GIGA-GAN & \cite{Kang_2023_CVPR} \\
    \bottomrule
  \end{tabular}
\end{table}

 VAE was also used for the molecule generations, generating the 3-Dimensional synthetic molecule structure \cite{zeng2022deep}. Jumper et al. \cite{jumper2021highly} developed an architecture of a generative algorithm specifically for predicting the structure of molecules and proteins. This is the best approach till now in generative AI techniques for generating and predicting molecular architectures.

\subsection{Generative AI for Video Synthesis and Generation}\label{subsec4.2}
Generative AI has transformative applications in the field of video and animation, enabling the creation of visually stunning and dynamic content.
The evaluation of video generative models often involves a set of widely recognized datasets:  Voxcleb \cite{Nagrani_2017vox}, HDTF \cite{zhang2021flow},  ClebV \cite{wu2018reenactganclebv}, Kinetics \cite{kay2017kinetics}, UCF101 \cite{soomro2012ucf101}, and for specifically audio performance testing VCTK Corpus\cite{oord2016wavenet} and LibriSpeech \cite{librispeech}. These datasets are publicly available to researchers, with the stipulation that any use of these resources must include proper citation of the source.

In Hong et al. \cite{Hong_2022_CVPR} the researchers introduced a GANs variant called depth-aware GAN. This model provided strong competition to the state of art models and the problem of replacing the face in a video. In detail the dataset on which they tested the built model and compared the results was called Voxcleb \cite{Nagrani_2017vox} and ClebV \cite{wu2018reenactganclebv}. The model produced better results than the present state-of-the-art models for achieving talking head video face generation.

Indeed, the replacement of a face in a talking head video is the most predominant application in the video generation problem. Many advancements have been made in generative Artificial intelligence to master this application. Hong et al. \cite{hong2023dagan} state that all this model needs is a target video and a 2-D picture with good pixels and facial features and it will translate the video expression features into the static picture supplied.
The \emph{DaGAN++} framework proposed by \cite{hong2023dagan} comprises three key components: (a) an uncertainty-aware face depth learning network that reconstructs detailed 3D facial geometry from self-supervised face videos, without requiring camera parameters or explicit 3D annotations (b) geometry-guided facial keypoint detection, which employs the facial depth network to estimate depth maps, is used alongside RGB images for accurate facial keypoint estimation (c) a geometry-enhanced multi-layer generation process that incorporates learned motion fields, occlusion maps, and facial geometry into each layer of image generation through cross-modal geometry-guided attention. This comprehensive approach enables the synthesis of images enriched with geometry-related attributes derived from facial videos.
 \begin{figure}[ht]
    \centering
    \includegraphics[width=0.7\textwidth]{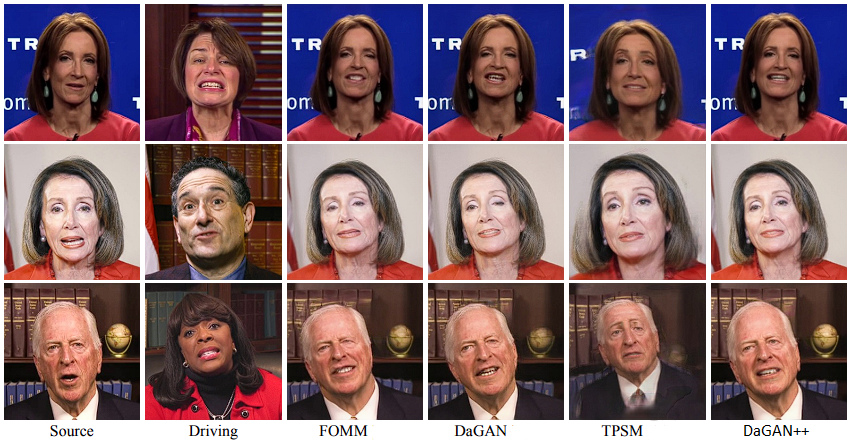}
     \caption{DaGAN++ vs other state of art model on HDTF dataset~\cite{zhang2021flow}, Source: \cite{hong2023dagan}}
     \label{dagan}
\end{figure}

It is evident from Figure \ref{dagan} that DaGAN++ exhibits superior capabilities. Notably, DaGAN++ excels in capturing expression-related facial movements within the driving frame, with heightened accuracy observed in regions such as the eyes and mouth. This performance enhancement can be attributed to the precise facial geometry estimation, which greatly contributes to the refinement of expression-related facial motions.
Furthermore, Min et al. \cite{min2022styletalker} introduce StyleTalker, an innovative audio-driven talking head generative model designed to synthesize a talking person's video using a single reference image. It features highly accurate lip synchronization, realistic head poses, and natural eye blinks synchronized to the provided audio. To achieve this, they leverage a pre-trained image generator and an image encoder to estimate latent codes for the talking head video that align faithfully with the given audio input. This achievement stems from the integration of novel components which includes a contrastive lip-sync discriminator. This ensures precise lip synchronization, a conditional sequential variational autoencoder that captures a motion space disentangled from lip movements. In more detail, it enables independent manipulation of motions and lip movements while preserving identity. In addition, it affords an auto-regressive prior enhanced with normalizing flow, facilitating the acquisition of a complex audio-to-motion multi-modal latent space. With these components in place, StyleTalker has the capacity to produce talking head videos, both in a motion-controllable manner when another motion source video is available. Also, entirely driven by audio inputs, wherein it infers real motions from the provided audio.
 \begin{figure}[ht]
    \centering
    \includegraphics[width=0.7\textwidth]{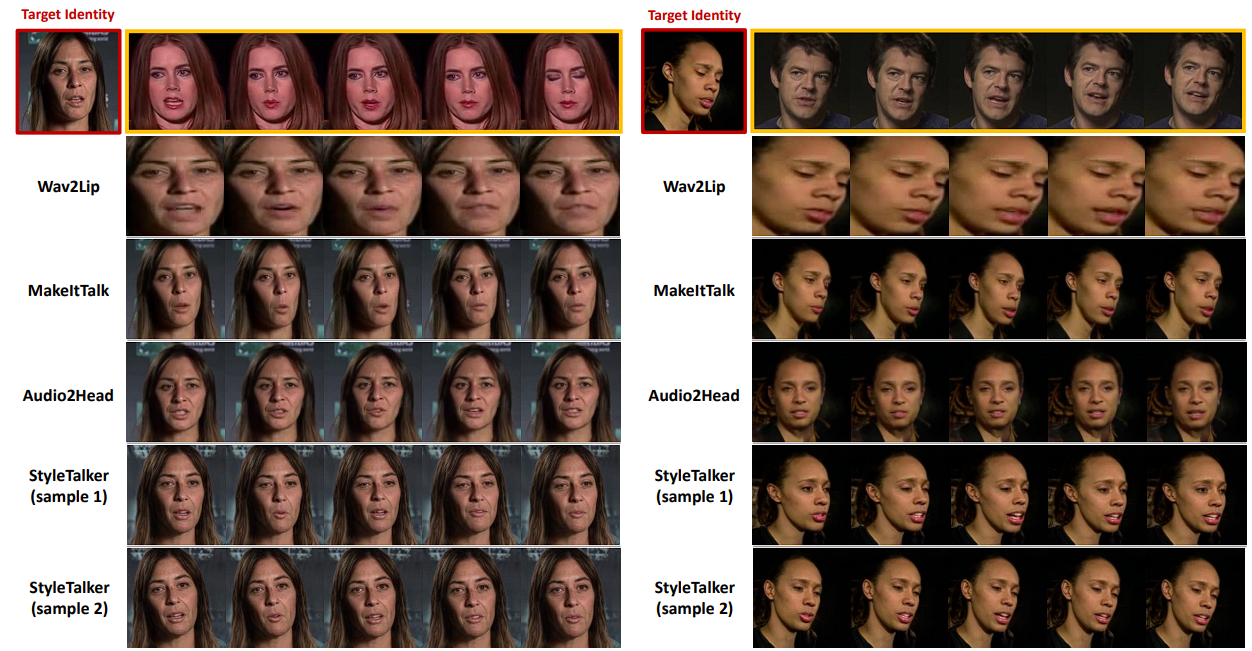}
     \caption{StyleTalker vs Other models, Source: \cite{min2022styletalker}}
     \label{styletalker}
\end{figure}

The qualitative evaluation of audio-driven talking head generation performance on VoxCeleb2 dataset in Figure \ref{styletalker} reveals distinct differences. In the first row (marked by a yellow box), frames corresponding to the provided audio are displayed. Conversely, the single image input (highlighted in a red box) represents a reference image of the desired target identity. When observing the generated video frames produced by the StyleTalker in comparison to those generated by other audio-driven generation models \cite{prajwal2020lip,zhou2020makelttalk,wang2021audio2head}, a notable distinction becomes evident. StyleTalker consistently produces talking head videos of exceptional quality, skillfully preserving the distinctive identity of the intended.

Advancing further, Li et al. \cite{Li_2022_CVPR} introduced Multiscale Vision Transformers (MViTv2) as a comprehensive architectural framework suitable for tasks encompassing image and video classification, along with object detection. Within this study, the researchers introduce an enhanced version of MViT, featuring decomposed relative positional embeddings and residual pooling connections. By deploying this upgraded architecture across five different scales, we meticulously assess its performance in scenarios such as ImageNet classification, COCO object detection, and Kinetics video recognition. Remarkably, MViTv2 surpasses previous benchmarks in terms of effectiveness and provided video classification accuracy of 86.1\% on the Kinetics-400 dataset as shown in Table \ref{Mvit}.
\begin{table}[htbp]
  \caption{Comparative analysis with other Models on the Kinetics-400 \cite{kay2017kinetics} dataset, Source: \cite{Li_2022_CVPR}}
  \centering

    \begin{tabular}{lcccc}
      \toprule
      \textbf{Model} & \textbf{Top-1} & \textbf{Top-5} & \textbf{FLOPs$\times$views} & \textbf{Param} \\
      \midrule
      SlowFast 16$\times$8 +NL \cite{feichtenhofer2019slowfast} & 79.8 & 93.9 & 234$\times$3$\times$10 & 59.9 \\
      X3D-XL \cite{feichtenhofer2020x3d} & 79.1 & 93.9 & 48.4$\times$3$\times$10 & 11.0 \\
      MoViNet-A6 \cite{kondratyuk2021movinets} & 81.5 & 95.3 & 386$\times$1$\times$1 & 31.4 \\
      MViTv1, 16$\times$4 \cite{fan2021christoph} & 78.4 & 93.5 & 70.3$\times$1$\times$5 & 36.6 \\
      MViTv1, 32$\times$3 \cite{fan2021christoph} & 80.2 & 94.4 & 170$\times$1$\times$5 & 36.6 \\
      \textbf{MViTv2-S}, 16$\times$4 \cite{Li_2022_CVPR} & \textbf{81.0} & \textbf{94.6} & 64$\times$1$\times$5 & 34.5 \\
      \textbf{MViTv2-B}, 32$\times$3 \cite{Li_2022_CVPR} & \textbf{82.9} & \textbf{95.7} & 225$\times$1$\times$5 & 51.2 \\
      ViT-B-VTN in 21k \cite{neimark2021video} & 78.6 & 93.7 & 4218$\times$1$\times$1 & 114.0 \\
      ViT-B-TimeSformer \cite{bertasius2021space} in 21k & 80.7 & 94.7 & 2380$\times$3$\times$1 & 121.4 \\
      ViT-L-ViViT \cite{arnab2021vivit} in 21k & 81.3 & 94.7 & 3992$\times$3$\times$4 & 310.8 \\
      Swin-L$^+$ in 21k \cite{liu2022video} & 84.9 & 96.7 & 2107$\times$5$\times$10 & 200.0 \\
      \textbf{MViTv2-L$^+$, 40$\times$3}, in 21k \cite{Li_2022_CVPR} & \textbf{86.1} & \textbf{97.0} & 2828$\times$3$\times$5 & 217.6 \\
      \bottomrule
    \end{tabular}

  \label{Mvit}
\end{table}

In the context of the Kinetics-400 dataset \cite{kay2017kinetics}, Table \ref{Mvit} presents a comparison between MViTv2 and previous methodologies, encompassing both state-of-the-art Convolutional Neural Networks (CNNs) and Vision Transformers (ViTs). Upon training from the ground up, MViTv2-S and MViTv2-B models exhibit top-1 accuracy of 81.0\%  and 82.9\% respectively, surpassing their MViTv1 \cite{fan2021christoph} counterparts by +2.6\% and +2.7\%. 
Notably, earlier ViT-based models necessitate substantial pre-training on the IN-21K dataset to achieve peak accuracy on Kinetics-400 as shown in Table \ref{Mvit} (see last 5 rows). In contrast, MViTv2 achieves an exceptional top-1 accuracy of 86.1\% when fine-tuning, the MViTv2-L model with a large spatio-temporal input size of $40 \times 3$ (time $\times$ space$^2$).

The rest of the reviewed applications of Generative Artificial Intelligence in the field of Video Generation are given in Table \ref{video generative table}

\begin{table}[h!]
  \centering
    \caption{Generative AI and Its Applications in Video Generation}
  \begin{tabular}{p{4cm}p{4cm}p{4cm}}
    \toprule
    Domain & Methods & References \\
    \midrule
    Face swapping videos & Depth Awareness GANs, DaGAN++ & \cite{Hong_2022_CVPR, hong2023dagan} \\
    Video/Image Classification & MViTv2 & \cite{Li_2022_CVPR} \\
    Audio-Based Facial Expression video Translation & Styletalker Sequential VAE & \cite{min2022styletalker} \\
    Simulation in Metaverses & Multi-task DT offloading model & \cite{xu2023generative} \\
    ECG Synthesis to Improve Deep ECG Classification & SimGAN & \cite{pmlr-v119-golany20a} \\
    3D human motion prediction & HP-GAN & \cite{Barsoum_2018_CVPR_Workshops} \\
    \bottomrule
  \end{tabular}

  \label{video generative table}
\end{table}

\subsection{Generative AI for Natural Language Processing}\label{subsec4.3}
Generative AI models have demonstrated remarkable achievements across a spectrum of Natural Language Processing tasks. These encompass language comprehension, logical reasoning, and text generation.

In the domain of natural language processing, specific datasets have become standard benchmarks for evaluating state-of-the-art models in various tasks. For Named Entity Recognition, the CoNLL-2003 dataset \cite{sang2003introduction} is frequently utilized. In the area of text summarization, two prominent datasets are DUC 2002 \cite{duc2002} and QMSUM \cite{zhong-etal-2021-qmsum}. Additionally, for Natural Language Inference (NLI) tasks across multiple languages, the XNLI dataset \cite{conneau2018xnli} serves as a crucial resource. All of these datasets are publicly available, providing researchers with essential tools to advance and assess the capabilities of their models in these specific NLP tasks.

Presently, a significant query the AI community poses revolves around the extent and confines of these model's capabilities \cite{ahuja2023mega}. Ahuja et al. \cite{ahuja2023mega} raise the question that most of the large language models are made and tested on only the English language. Therefore these researchers took the state-of-the-art models and trained them on other languages on certain available datasets and compared their question-answering and classificational accuracies \cite{ahuja2023mega}. Generative AI even has found its way into education\cite{cao2023comprehensive,DWIVEDI2023102642}. In the midst of this dynamic backdrop that challenges conventional modes of thinking, recent research endeavors investigating the implications of generative AI within the educational landscape yield valuable insights. Notably, these studies shed light on the opportunities and obstacles arising from the integration of generative AI. For instance, in a recent article \cite{tlili2023if} 
highlights the need for a new and creative way of teaching that effectively incorporates the progress brought by AI. They mentioned the significance of cultivating an ethical and personalized chatbot solution while augmenting digital proficiency to fully harness the manifold benefits of AI. Furthermore, the researchers advocate for the incorporation of AI literacy as an essential technological skill for navigating the complexities of the 21st century.

In the current landscape, Bozkurt \cite{bozkurt2023generative} advocates for a significant reevaluation of the roles played by human educators and AI within the educational realm. They assert that the emergence of AI offers a unique juncture to redefine these roles. Especially, as AI possesses the capacity to assume an increasing array of educational tasks that were traditionally the exclusive domain of human educators. This perspective underscores the importance of adopting a forward-thinking outlook that reconsiders the contributions of both technology and human educators to the educational process. Bozkurt \cite{bozkurt2023generative} also emphasizes that generative AI's arrival presents a propitious opportunity to redefine these roles further.

The researchers further delve into the opportunities and challenges ushered in by the advent of generative AI. Generative AI, they elaborate, provides a diverse range of opportunities. These encompass personalized learning, fostering inclusive curriculum provision, and enhancing collaboration and cooperation throughout educational processes. Also, they can cater to automated assessment benefits, ensuring improved accessibility, optimizing efficiency in terms of time and effort, cultivating language skills, and enabling the round-the-clock availability of these technologies. 

Generative AI can also enhance synthetic data generation, with the use of Transformers and GANs. It is clear that the understanding of data is better and now the AI is able to generate synthetic data even in the health field. For example, Frid-Adar et al. \cite{fridadar2018synthetic} used GANs to generate synthetic data for the liver lesion classification problem. The variant they used was called DCGAN. In another problem, Wang et al. \cite{wang2019learning} used the SSIM embedded-cycle GAN to count the number of people in the crowd. The dataset that they used for this problem was fully synthetically generated. It had all the weather conditions covered which made the model perform the best compared to the state-of-the-art models. 
Another technique of generative AI that is among the state-of-the-art techniques is BERT (Bidirectional Encoder representations from Transformers) which was introduced by the Google AI team \cite{devlin2018bert}. It represents a significant advancement in pre-training techniques for NLP tasks. BERT is based on the transformer architecture and is designed to capture contextual information from both the left and right sides of a word in a sentence, hence the term ‘bidirectional’.
In Bert's pre-training, the model learns to predict missing words in a sentence by training on a large corpus of text. This helps BERT develop a deep understanding of syntax, semantics, and context. After pre-training, the model is fine-tuned on specific downstream tasks, such as sentiment analysis, question answering, and named entity recognition, using task-specific labelled data.
\begin{table}[h!]
  \centering
   \caption{ BERT vs others for Named Entity Recognition task on the CoNLL-2003~\cite{sang2003introduction} dataset (comparison in terms of F1 Score on Validation DataSet (Dev) and Testing Dataset (Test), Source:\cite{devlin2018bert}}
  \begin{tabular}{p{4cm}cc}
    \toprule
    \textbf{System} & \textbf{Dev F1} & \textbf{Test F1} \\
    \midrule
    ELMo \cite{joshi2018extending} & 95.7 & 92.2 \\
    CVT \cite{clark2018semi} & - & 92.6 \\
    CSE \cite{akbik2018contextual} & - & 93.1 \\
    \midrule
    \textbf{Fine-tuning approach} & & \\
    BERTLARGE & 96.6 & 92.8 \\
    BERTBASE & 96.4 & 92.4 \\
    \midrule
    \textbf{Feature-based approach (BERTBASE)} & & \\
    Embeddings & 91.0 & - \\
    Second-to-Last Hidden & 95.6 & - \\
    Last Hidden & 94.9 & - \\
    Weighted Sum Last Four Hidden & 95.9 & - \\
    Concat Last Four Hidden & 96.1 & - \\
    Weighted Sum All 12 Layers & 95.5 & - \\
    \bottomrule
  \end{tabular}

\label{bertresult}
\end{table}

Table~\ref{bertresult} shows the BERT model tested with the CoNLL-2003~\cite{sang2003introduction} dataset for the task focusing on named entity recognition. BERTLarge demonstrates strong competitiveness with state-of-the-art techniques. The most successful approach involves concatenating token representations from the uppermost four hidden layers of the pre-trained Transformer. Remarkably, this approach lags by only 0.3 F1 behind the performance achieved by fine-tuning the complete model. This finding underscores the effectiveness of BERT for both fine-tuning and feature-based methodologies.

Another model related to Generative Natural language processing is ELMo~\cite{joshi2018extending}. This stands for ‘Embeddings from Language Models'. ELMo utilizes a bidirectional LSTM (Long Short-Term Memory) network \cite{hochreiter1996lstm} for contextual word embeddings. In fact, ELMo embeddings have been shown to be effective in improving the performance of various NLP tasks, including sentiment analysis, question answering, and named entity recognition. The ability to capture context-specific information makes ELMo embeddings particularly useful for tasks where word meanings can vary based on the surrounding context.

Another application of generative AI models used in natural language processing is malware classification. In particular, machine language malware classification is a big concern that can be solved by using generative AI. For example, Kale et al. \cite{kale2023malware}  used the Bert and ELMo to train the embeddings of the models to classify the malware and the results provided remarkable improvements.
 \begin{figure}[ht]
    \centering
    \includegraphics[width=0.7\textwidth]{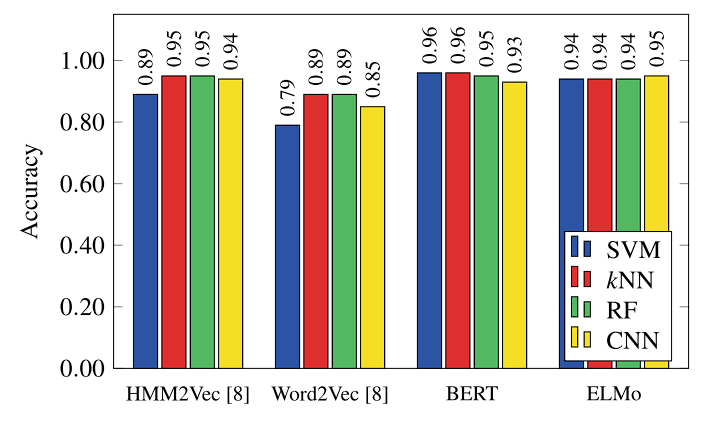}
     \caption{HMM2Vec, Word2Vec, BERT, and ELMo based classification techniques, Source: \cite{kale2023malware}}
     \label{emlomalware}
\end{figure}

Additionally, Figure \ref{emlomalware} provides a summarized depiction of the optimal accuracies achieved by the classification methodologies based on HMM2Vec\cite{chandak2021comparison}, Word2Vec\cite{mikolov2013efficient}, BERT\cite{devlin2018bert}, and ELMo\cite{peters2018deep}. The graph illustrates that BERT-SVM and BERT-kNN secured the highest performance with a notable accuracy of 96\%. In close pursuit HMM2Vec-kNN, HMM2Vec-RF, BERT-RF, and ELMo-CNN, achieved an accuracy of 95\%. Slightly trailing behind, HMM2Vec-CNN, as well as all ELMo-based techniques, demonstrated an accuracy of 94\%. Interestingly, the Word2Vec embeddings consistently yielded accuracies below 90\% across all four classifiers.

A further application of Generative artificial intelligence is the text-summarization. Joshi et al. \cite{joshi2022ranksum} introduced \emph{Ranksum} which is an innovative technique designed for extractive text summarization of individual documents. This method hinges on the fusion of four distinct multi-dimensional sentence features, namely topic information, semantic content, significant keywords, and position. By independently acquiring sentence saliency rankings for each feature in an unsupervised manner, Ranksum subsequently amalgamates these scores through weighted fusion, yielding a comprehensive ranking of sentence significance. It is important to note that these scores are generated in a completely unsupervised manner.

Topic ranking is established through the application of probabilistic topic models, while semantic content is captured using sentence embeddings. Sentence embeddings are generated using Siamese networks to craft abstractive sentence representations, followed by a novel strategy to organize them based on their relative importance. To identify significant keywords and their associated sentence rankings within the document, a graph-based approach is employed. Additionally, a mechanism to gauge sentence novelty is formulated, relying on bigrams, trigrams, and sentence embeddings. This eliminates redundant sentences from the summary. 
\begin{table}[h!]
  \centering
    \caption{Comparative analysis of RankSum with state-of-the-art algorithms conducted on the DUC 2002 \cite{duc2002} and QMSUM \cite{zhong-etal-2021-qmsum} dataset, Source: \cite{joshi2022ranksum, yang2023exploring}}
  \begin{tabular}{lccc}
    \toprule
    \textbf{Method} & \textbf{ROUGE-1} & \textbf{ROUGE-2} & \textbf{ROUGE-L} \\
    \midrule
    LEAD & 43.6 & 21.0 & 40.2 \\
    ILP & 45.4 & 21.3 & 42.8 \\
    NN-SE & 47.4 & 23.0 & – \\
    SummaRuNNer & 47.4 & 24.0 & 14.7 \\
    Egraph+coh & 47.9 & 23.8 & – \\
    Tgraph+coh & 48.1 & 24.3 & – \\
    URANK & 48.5 & 21.5 & – \\
    SummCoder & 51.7 & 27.5 & 44.6 \\
    \textbf{HSSAS} & \textbf{52.1} & 24.5 & \textbf{48.8} \\
    CoRank & 52.6 & 25.8 & – \\
    Rank-emb & 49.9 & 24.8 & 45.6 \\
    Rank-topic & 51.4 & 25.9 & 47.2 \\
    Rank-keyword & 52.0 & 26.3 & 48.6 \\
    \textbf{RankSum} & \textbf{53.2} & 27.9 & \textbf{49.3} \\
    PGNet on QMSUM & 31.52 & 8.69 & 27.63 \\
    BART on QMSUM & 32.18 & 8.48 & 28.56 \\
    \textbf{HMNet} on QMSUM & \textbf{36.06} & \textbf{11.36} & \textbf{31.27} \\
    \textbf{ChatGPT} on QMSUM & \textbf{36.83} & \textbf{12.78} & \textbf{24.23} \\
    \bottomrule
  \end{tabular}

  \label{ranksum}
\end{table}

Table \ref{ranksum} showcases the performance outcomes of the novel RankSum framework in comparison to other state-of-the-art algorithms on the DUC 2002 dataset, evaluated through ROUGE metrics. ROUGE metrics are typically used in the field of machine translation, text summarization, and other tasks where the quality of generated text needs to be evaluated automatically. The metrics involve comparing n-grams (sequences of n words) between the generated and reference texts. Common versions of ROUGE metrics include ROUGE-N (which considers overlapping n-grams), ROUGE-L (which focuses on the longest common subsequence) \cite{lin-2004-rouge}. Ranksum achieves notable ROUGE-1, ROUGE-2, and ROUGE-L scores of 53.2, 27.9, and 49.3, respectively. Impressively, it outperforms all recent methods examined for this extractive text summarization dataset. Notably, this approach surpasses the highly accurate summarization systems, HSSAS \cite{al2018hierarchical} and Co-Rank, with a substantial margin of 0.6, 0.8, and 0.5 for ROUGE-1, ROUGE-2, and ROUGE-L scores respectively \cite{joshi2022ranksum}. Additionally, the outcomes of PGnet \cite{wang2021pgnet}, BART\cite{lewis2019bart}, HMNet\cite{xiao2021hmnet}, and ChatGPT\cite{brown2020language} were assessed using the QMSUM dataset\cite{zhong-etal-2021-qmsum}. It is important to note that the current pinnacle of meeting summarization models, HMNet, achieves the most impressive performance in terms of ROUGE-L. This might be attributed to its cross-domain pretraining approach, which imparts HMNet with a heightened familiarity with the style of meeting transcripts \cite{zhu-etal-2020-hierarchical}. However, it's worth highlighting that in the case of ROUGE-1 and ROUGE-2 metrics, ChatGPT emerges as the leader. ChatGPT excels due to extensive training on diverse data, enabling superior relationship comprehension in one to two grams, and boosting metric performance.
\begin{table}[htbp]

  \centering

  \resizebox{\columnwidth}{!}{
    \begin{tabular}{lcccccccccccccccc}
      \toprule
      \textbf{Model} & \textbf{en} & \textbf{ar} & \textbf{bg} & \textbf{de} & \textbf{el} & \textbf{es} & \textbf{fr} & \textbf{hi} & \textbf{ru} & \textbf{sw} & \textbf{th} & \textbf{tr} & \textbf{ur} & \textbf{vi} & \textbf{zh} & \textbf{avg} \\
      \midrule
      Fine-tuned Baselines & 80.8 & 64.3 & 68.0 & 70.0 & 65.3 & 73.5 & 73.4 & 58.9 & 67.8 & 49.7 & 54.1 & 60.9 & 57.2 & 69.3 & 67.8 & 65.4 \\
      Prompt-Based Baselines & 67.5 & 60.7 & 46.5 & 54.0 & 47.4 & 61.2 & 61.4 & 56.8 & 53.3 & 50.4 & 43.8 & 42.7 & 50.0 & 61.0 & 56.7 & 54.2 \\
      Open AI Models & 76.2 & 59.0 & 63.5 & 67.3 & 65.1 & 70.3 & 67.7 & 55.5 & 62.5 & 56.3 & 54.0 & 62.6 & 49.1 & 60.9 & 62.1 & 62.1 \\
      mBERT & 80.8 & 64.3 & 68.0 & 70.0 & 65.3 & 73.5 & 73.4 & 58.9 & 67.8 & 49.7 & 54.1 & 60.9 & 57.2 & 69.3 & 67.8 & 65.4 \\
      mT5-Base & 84.7 & 73.3 & 78.6 & 77.4 & 77.1 & 80.3 & 79.1 & 70.8 & 77.1 & 69.4 & 73.2 & 72.8 & 68.3 & 74.2 & 74.1 & 75.4 \\
      XLM-R Large & 88.7 & 77.2 & 83.0 & 82.5 & 80.8 & 83.7 & 82.2 & 75.6 & 79.1 & 71.2 & 77.4 & 78.0 & 71.7 & 79.3 & 78.2 & 79.2 \\
      TuLRv6 - XXL & 93.3 & 89.0 & 90.6 & 90.0 & 90.2 & 91.1 & 90.7 & 86.2 & 89.2 & 85.5 & 87.5 & 88.4 & 82.7 & 89.0 & 88.4 & 88.8 \\
      gpt-3.5-turbo & 76.2 & 59.0 & 63.5 & 67.3 & 65.1 & 70.3 & 67.7 & 55.5 & 62.5 & 56.3 & 54.0 & 62.6 & 49.1 & 60.9 & 62.1 & 62.1 \\
      gpt-3.5-turbo (TT) & 76.2 & 62.7 & 67.3 & 69.4 & 67.2 & 69.6 & 69.0 & 59.9 & 63.7 & 55.8 & 59.6 & 63.8 & 54.0 & 63.9 & 62.6 & 64.3 \\
      text-davinci-003 & 79.5 & 52.2 & 61.8 & 65.8 & 59.7 & 71.0 & 65.7 & 47.6 & 62.2 & 50.2 & 51.1 & 57.9 & 50.0 & 56.4 & 58.0 & 59.3 \\
      text-davinci-003 (TT) & 79.5 & 65.1 & 70.8 & 71.7 & 69.3 & 72.2 & 71.8 & 63.3 & 67.3 & 57.3 & 62.0 & 67.6 & 55.1 & 66.9 & 65.8 & 67.1 \\
      gpt-4-32k & 84.9 & 73.1 & 77.3 & 78.8 & 79.0 & 78.8 & 79.5 & 72.0 & 74.3 & 70.9 & 68.8 & 76.3 & 68.1 & 74.3 & 74.6 & 75.4 \\
      \bottomrule
    \end{tabular}
    }
  \caption{Performance comparison among different models on all languages within the XNLI dataset, Source:\cite{ahuja2023mega}.}
  \label{multilingual}
\end{table}

Ahuja et al. \cite{ahuja2023mega} tested the performance of the state-of-the-art models on multilingual XNLI dataset \cite{conneau2018xnli} data. The results are given in Table \ref{multilingual}, TuLRv6 - XXL achieves the highest average accuracy across all languages (88.8\%). It performs exceptionally well in most languages, with accuracy scores consistently above 85\%, XLM-R Large is the model that comes in second place with an average accuracy of 79.2\%. While not quite as high as TuLRv6, it still maintains a strong performance across all languages and demonstrates its multilingual capabilities. With an average accuracy of 75.4\%, mT5-Base takes the third spot. GPT-4-32k achieves an average accuracy of 75.4\%. It exhibits consistent performance across languages, demonstrating its effectiveness in handling multilingual tasks.

\begin{longtable}{p{3.5cm}p{3cm}p{5cm}p{1cm}}
\caption{Generative AI Applications in Natural Language Processing: Major Papers and Descriptions} \label{NLPTimeline} \\
    \hline
    \textbf{Paper} & \textbf{Field} & \textbf{Description} & \textbf{Citation} \\
    \hline
    \endfirsthead
    
    \multicolumn{4}{c}{{\tablename\ \thetable{} -- continued from previous page}} \\
    \hline
    \textbf{Paper} & \textbf{Field} & \textbf{Description} & \textbf{Citation} \\
    \hline
    \endhead
    
    \hline 
    \multicolumn{4}{r}{{Continued on next page}} \\
    \endfoot
    
    \hline
    \endlastfoot

    ``Attention Is All You Need" & NLP / Machine Translation & Introduces the Transformer model using self-attention mechanisms for various NLP tasks, revolutionizing sequence-to-sequence models & \cite{NIPS2017_attention}\\
    \hline
   ``BART: Denoising Sequence-to-Sequence Pre-training for Natural Language Generation, Translation, and Comprehension"&NLP / Text Generation&Introduces BART, a sequence-to-sequence model pre-trained using denoising autoencoders, capable of various NLP tasks& \cite{lewis2019bart}\\
    \hline
    ``CTRL: A Conditional Transformer Language Model"&Language Generation&Proposes a generative model (CTRL) that can condition its output on specific attributes, enabling fine-grained control over text generation& \cite{keskar2019ctrl}\\
    \hline
    ``T5: Exploring the Limits of Transfer Learning with a Unified Text-to-Text Transformer"&NLP / Transfer Learning&Presents T5, a model that casts all NLP tasks as a text-to-text problem, achieving state-of-the-art results across diverse tasks& \cite{raffel2020exploring}\\
    \hline
    ``GPT-2: Language Models are Unsupervised Multitask Learners" & Language Generation&Describes the GPT-2 model, a large-scale generative model that demonstrates impressive text generation capabilities across a range of tasks& \cite{radford2019language}\\
    \hline
    ``LayoutLM: Pre-training of Text and Layout for Document Image Understanding"&Document Analysis&Presents LayoutLM, a model that pre-trains on document images with associated text, improving document understanding tasks&\cite{Xu_2020}\\
    \hline
    ``ERNIE: Enhanced Language Representation with Informative Entities"&NLP / Knowledge Enhancement&Introduces ERNIE, a model that enhances language representations by incorporating knowledge from knowledge bases& \cite{zhang2019ernie}\\
    \hline
    ``DALL·E: Creating Images from Text"&NLP / Image Generation&Introduces DALL·E, a generative model capable of generating images from textual descriptions& \cite{reddy2021dall}\\
    \hline
    ``CLIP: Connecting Text and Images for Supervised Learning"&NLP / Vision&Proposes CLIP, a model that learns to understand images and text jointly, achieving impressive results in cross-modal tasks& \cite{radford2021learning}\\
    \hline
    ``WebGPT: Browser-assisted question-answering with human feedback" & NLP, Human feedback & Introduces an approach that leverages a text-based web-browsing environment, enabling the model to access and navigate online resources. The methodology is structured in a manner that aligns with human capabilities, thus facilitating model training through imitation learning.& \cite{nakano2022webgpt}\\
    \hline
    ``GPT-4 Technical Report" & NLP/ Text Generation &GPT-4, a groundbreaking advancement, is introduced as a multimodal model with the ability to process both image and text inputs while generating text-based outputs. GPT-4's accomplishments encompass passing a simulated bar exam with a score that ranks within the top 10\% of test takers.  & \cite{openai2023gpt4}\\
    \hline
    ``Let's Verify Step by Step" & Mathematical Reasoning & The introduced approach, centered around a process-supervised model, achieves a commendable success rate of 78\% when addressing problems sourced from a representative subset of the MATH test set. & \cite{lightman2023lets}\\
    \hline
\label{NLPTimeline}
\end{longtable}

The landscape of Natural Language Processing (NLP) has witnessed remarkable advancements in recent years, driven primarily by the innovative applications of generative AI. The Table \ref{NLPTimeline} highlights a selection of influential papers that showcase the evolution and impact of generative AI techniques within the NLP domain. These advancements have led to groundbreaking developments in various subfields of NLP, transforming the way we process and understand human language.

As evidenced by the papers presented, recent years have seen generative AI techniques reshape NLP in profound ways. These advancements not only enhance the quality and diversity of text generation but also enable more sophisticated control, cross-modal understanding, and knowledge integration. As the field continues to evolve, it is likely that further innovations in generative AI will continue to drive NLP's progress, unlocking new frontiers of language understanding and generation across a wide array of applications.

\subsection{Generative AI for Knowledge Graph Generation}\label{subsec4.4}
Researchers and practitioners have leveraged the power of generative AI to enhance the creation and refinement of knowledge graphs—a structured representation of relationships between entities.
This section explores the burgeoning landscape of generative AI applications within knowledge graph generation, highlighting pioneering research papers and their contributions to this evolving field.
For evaluating models in the field of knowledge graph generation, a variety of datasets are employed, with Dbpedia \cite{lehmann2015dbpedia}, Cora Dataset \cite{cabanes2012cora} and Googles Knowledge graph \cite{Singhal_2012} being among the most commonly used. However, the scope of available datasets extends beyond just these. These datasets are freely accessible, offering researchers and developers an invaluable resource to test and refine the capabilities of their knowledge graph generation models. 
A knowledge graph is a structured representation of information that captures relationships between entities and concepts. It goes beyond traditional databases by not only storing data but also organizing it in a way that highlights connections and context. Knowledge graphs, first introduced by Google in 2012 \cite{steiner2012adding}, are designed to model real-world relationships, making them a powerful tool for representing and querying complex information. 
Cai and Wang \cite{cai2018kbgan} introduced KBGAN, an innovative adversarial learning framework designed to enhance the performance of various existing knowledge graph embedding models. This approach is not dependent on the specific structures of the generator and discriminator, allowing for the incorporation of a wide range of knowledge graph embedding models as fundamental components. This enables KBGAN to significantly enhance the training dynamics and performance of existing knowledge graph embedding models.
Liu et al. \cite{liu2019kbert} introduce K-BERT, a novel approach that empowers language representation with knowledge graphs, enabling the incorporation of commonsense and domain-specific knowledge. The K-BERT methodology comprises two fundamental steps. Initially, knowledge from a knowledge graph (KG) is seamlessly integrated into a sentence, rendering it a knowledge-rich sentence tree. Subsequently, the utilization of soft-position and visible matrix techniques serves to regulate the extent of knowledge integration, thereby preventing any deviation from the original sentence meaning.

Despite the challenges presented by handling heterogeneous entity spans (HES) and keyphrases not seen in training (KN), the investigation yields promising outcomes across a spectrum of twelve open-domain and specific-domain natural language processing (NLP) tasks. Empirical evidence underscores the considerable efficacy of knowledge graphs, particularly in tasks that are driven by domain-specific knowledge. Moreover, K-BERT's compatibility with the model parameters of BERT offers a seamless integration of knowledge enhancement within a well-established framework.

Link prediction is a fundamental task involving the prediction of missing facts within a knowledge graph using available information. In this context, Balazevic et al. \cite{Balazevic_2019} introduce TuckER, a linear model that employs Tucker decomposition of the binary tensor representation of knowledge graph triples. Despite its straightforward nature, TuckER demonstrates remarkable efficacy. It surpasses previous state-of-the-art models on widely recognized link prediction datasets, solidifying its position as a potent baseline for more sophisticated models in this domain.
Moreover, Zeb et al. \cite{zeb2022complex} have introduced ComplexGCN, an innovative graph convolutional network that leverages standard graph convolutional architecture(GCN) \cite{kipf2017semisupervised} to learn complex embeddings. Within the ComplexGCN framework, both node and relation features are projected into complex space through the use of learnable weights associated with neighboring nodes at each convolutional layer. To maintain the integrity of initial embedding information in the final node embeddings, a residual connection between the input and output of the convolutional stack is implemented.
\begin{figure}[ht]
    \centering
    \includegraphics[width=0.7\textwidth]{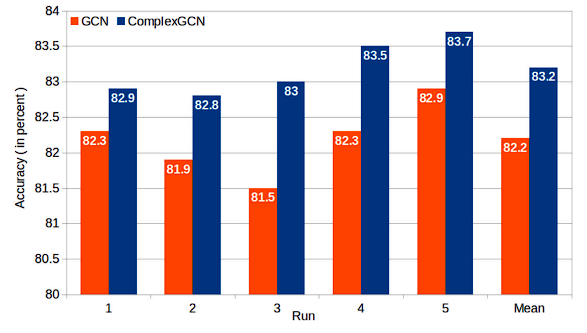}
     \caption{GCN vs CGCN on node completion task- Cora Dataset, Source: \cite{zeb2022complex}}
     \label{cgcn}
\end{figure}
These researchers \cite{zeb2022complex} conducted an evaluation of ComplexGCN's performance on the node classification task using the Cora dataset. This consisted of 2708 nodes categorized into 7 classes and 5429 edges representing citation links between documents. Both the standard GCN and ComplexGCN were trained on the Cora dataset for 200 epochs, with the objective of minimizing the cross-entropy loss. The training process was repeated 5 times for each model, and the results were averaged and reported in Figure \ref{cgcn}. In terms of accuracy percentage, ComplexGCN exhibited improved performance compared to the standard GCN, achieving a 1\% increase in mean accuracy.

The presented selection of papers highlights the innovative strides researchers have made in leveraging generative AI to enhance knowledge graph generation. From employing adversarial learning for improved embeddings to bridging the gap between unstructured text and structured knowledge, these papers showcase the multifaceted nature of the advancements.
\subsection{Interdisciplinary Applications of Generative AI}\label{subsec4.5}
The groundbreaking capabilities of generating synthetic data and content generation have given Generative AI the ability to be applicable in interdisciplinary fields. Further recent applications are discussed in this section.
The synthetic data generation capability of generative AI is useful in mechanical fault detections. Gao et al. \cite{fem} highlight a fault detection method that combined Finite Element Method (FEM)\cite{courant1943variational} simulations and Generative Adversarial Network (GAN) \cite{goodfellow2014generative} to address two primary challenges. Firstly, it aimed to fill the gaps in fault samples by leveraging FEM simulations. Secondly, it sought to enhance fault detection accuracy by utilizing GAN to generate a substantial number of synthetic fault samples. Initially, FEM was employed to generate simulation signals of specific lengths to complete missing fault samples. Subsequently, GAN was utilized to create additional fault samples based on FEM simulations, resulting in a more comprehensive dataset. Finally, classifiers such as Support Vector Machines (SVM)\cite{cortes1995support}, Extreme Learning Machines (ELM)\cite{huang2004extreme}, Decision Trees (DTree)\cite{quinlan1986induction}, and others were employed to detect faults in cases where the faults were previously unknown.
\begin{table}[h!]
\centering
\caption{Comparison of Classification accuracy with and without GANs sampling, Source: \cite{fem}}
\begin{tabular}{|l|c|c|}
\hline
\textbf{Description} & \textbf{FEM \& AI} & \textbf{FEM, GAN \& AI} \\
\hline
No of fault Samples & 360 & 360 \\
No of Synthetic Fault Samples & -- & 3240 \\
\hline
Accuracy SVM (\%) & 84.44 & 86.11 \\
Accuracy ELM (\%) & 89.72 & 91.67 \\
Accuracy Decision Tree (\%) & 91.11 & 97.5 \\
Average Accuracy (\%) & 88.42 & 91.76 \\
\hline
\end{tabular}

\label{bearingfault}
\end{table}

In Table \ref{bearingfault} the result clearly identifies the use of Generative AI to generate the synthetic samples which helps the machine learning models to achieve greater accuracy.
In Feng et al. \cite{traffic2023} a traffic generation model is developed using the Generative AI, named TrafficGen. The model outperforms the previous state-of-art techniques such as SeneGen \cite{senegen} in generating the synthetic traffic scenarios. The introduced models were also able to generate the trajectory of the generated traffic and synthetic snapshots. This enabled the creation of numerous fresh traffic scenarios and the enhancement of the ones that already exist.

Music generation is also gaining popularity in the field of Generative AI, different methods such as Variational Auto Encoders, Transformers and Recurrent Neural Networks are being used in generating synthetic music, and different new approaches of music generation  \cite{musicreview}. Another application is the Handwriting generation \cite{handwriting}. This proposed HiGAN+ which presented the capability to generate a wide range of authentic handwritten texts while being guided by arbitrary textual content and distinct calligraphic styles. These styles are separated from reference images or randomly drawn from a prior normal distribution. Traditional style transfer methods, which rely on pixel-level mappings, may not be suitable for HiGAN+, hence they introduce the contextual loss to notably enhance the stylistic consistency of generated images. The model performed very well in generating readable handwriting samples.
In the field of software engineering, generative AI is introduced to help write better code and solve errors in the code, debugging and even writing the documentation of the work (e.g., Copilot) \cite{copilot}.

Emerging technologies in the realm of generative AI aim to simplify human life, yet they also underscore the imperative for responsible AI development. These innovations should prioritize ethical considerations, ensuring that their generated content aligns with societal values. The Contemporary efforts in the generative AI method development are increasingly conscious of these ethical dimensions.
In a recent review, Pudari and Ernst \cite{pudari2023copilot} delved into Copilot and emphasized that generative AI, while valuable, won't supplant humans in the field of software engineering. This is because it struggles to grasp intricate software design principles and identify coding issues, known as ‘code smells’. Instead, its role primarily revolves around aiding developers in crafting more efficient code. Researchers \cite{arrieta2020explainable} highlight responsible AI  as a comprehensive concept, mandating systematic adoption of AI principles. Besides explainability, it emphasizes fairness, accountability, and privacy in real-world AI model implementations, especially in scenarios involving sensitive information and regulatory demands for data privacy.

Apart from these discussed papers, a multitude of groundbreaking advancements are unfolding within the domain of generative AI. It is evident that in this rapidly advancing research, the previously discussed papers represent just a prominent subset of the recent developments in this field.
\section{Challenges and opportunities of Generative AI}\label{sec5}
There are various domains for which we can discuss both the challenges and opportunities of Generative AI. Let’s start with challenges and their proposed solutions followed by opportunities- 
\subsection{Challenges and their proposed solutions}
\textbf{Ethical Concerns:}\\
Challenge: GenAI can be used for malicious purposes, such as the creation of deepfakes for identity theft or misinformation.\\
Solution: Establishing ethical governance structures, guidelines, and regulations to guide the responsible development and deployment of GenAI.\\
\textbf{Security Concerns:}\\
Challenge: There might be vulnerabilities in generative models that could be exploited for adversarial attacks.\\
Solution: The development of security measures to protect generative models from manipulation and continuous research into adversarial robustness.\\
\textbf{Bias and Fairness:}\\
Challenge: Generative models may amplify and perpetuate biases in the training data, leading to discriminatory and unfair outputs.\\
Solution: Extensive research and implementation of methods to detect and mitigate bias in training data, as well as encouraging inclusivity and diversity in datasets.\\
\textbf{Data Privacy:}\\
Challenge: Generative models trained on large datasets may inadvertently remember sensitive information, posing privacy risks.\\
Solution: Adherence to data protection regulations and implementation of privacy-preserving approaches to protect personal privacy.\\
\textbf{Interpretability:}\\
Challenge: Mostly, it is difficult to understand the decision-making process of generative algorithms due to their `black boxes' nature.\\
Solution: Research and development of explainable AI approaches to improve interpretability and transparency and, permitting users to understand algorithm outputs.
\subsection{Opportunities}
\textbf{Human-AI Collaboration:} Collaborative work between GenAI and humans can lead to innovative solutions in design, problem-solving, and creativity.\\
\textbf{Creative Expression:} Generative AI facilitates innovative creative expressions, such as music, generative art, and literature.\\
\textbf{Content Generation:} Applications in content creation, for example- image synthesis, text generation, and video creation enhance productivity in numerous industries.\\
\textbf{Education and Training:} Generative models can be employed for simulating scenarios for training, creating interactive educational materials, and enhancing learning experiences.\\
\textbf{Personalization:} Generative algorithms can be used for personalized recommendations in entertainment, e-commerce, and other user-oriented domains.\\
\textbf{Innovative Design:} GenAI can assist in creating optimized and innovative designs in industries like architecture and product design.\\
\textbf{Scientific Discovery: }GenAI contributes to scientific investigation by simulating complex systems, predicting outcomes, and generating hypotheses.\\
\textbf{Healthcare Applications:} GenAI advances personalized medicine, drug discovery, medical imaging, and healthcare system.\\
Balancing the potential benefits of generative AI with the demand for responsible development and deployment is essential. Ethical considerations, transparency, and ongoing research will play major roles in maximizing the positive impact of generative AI while minimizing risks.

\section{Conclusion and Future Direction}\label{sec6}
This paper offers a comprehensive systematic literature review of recent advancements in the field of generative AI. Specifically, it thoroughly explores key algorithms within the realm of Generative AI, including Diffusion Models, Transformer-based models, Generative Adversarial Networks, Variational Autoencoders, and their advancements tailored to specific applications.

Within the paper, we discuss advanced methodologies developed by various researchers, representing the current state-of-the-art achievements in the field of generative AI. A primary focus of generative AI's impact is evident in the domains of NLP and Video Translation, where advanced models have emerged with the capacity to tackle a wide array of human-centric challenges. These include tasks like question answering, code generation, language translation, image transformation, and more interdisciplinary applications. The paper highlights the recent achievements made in these areas, shedding light on the cutting-edge advancements achieved through generative AI techniques.

Moreover, it seems evident that the future direction of generative AI will be a transformative journey. One critical avenue of exploration involves the continuous evolution of AI architectures, aiming to create models that surpass current machine and human capabilities. Additionally, the ethical dimension of AI is set to gain even more prominence, with research and development focusing on ensuring responsible AI generation, minimizing biases, and aligning with evolving ethical standards. Indeed, interdisciplinary collaborations will flourish, as generative AI is applied to complex challenges in fields like healthcare, climate science, and education, amplifying its real-world impact. 

No doubt, the synergy between humans and AI will deepen, emphasizing AI's role as a collaborative partner across various domains. Advancements in NLP will persist, with an emphasis on question-answering, multilingual translation, and code generation. The domain of image, video, and multimedia processing will witness expansion, with generative AI contributing to content creation, enhancement, and interpretation. As we journey into this new and exciting future, it is also clear that we need to remain committed to responsible AI development and ethical considerations in parallel to developing these more advanced generative AI methods.

\section*{Statements and Declarations}

\subsection*{Funding}
This research received partial financial support from the Wales Innovation Network and Global Wales Small Grant Fund, Grant Number: GW-230433/414 (1.3.1b).

\subsection*{Conflict of interest/Competing interests}
All authors declare that they have no conflict of interest.

\subsection*{Ethics approval}
Not Required.

\subsection*{Data availability statement}
This research paper is a systematic review that does not utilize any specific dataset for analysis. Instead, it focuses on collating and evaluating existing literature to provide a comprehensive overview of the field.

\bibliographystyle{apalike}
\bibliography{sn-bibliography}
\end{document}